\documentclass[10pt,twocolumn,letterpaper]{article}

\usepackage{iccv}
\usepackage{times}
\usepackage{epsfig}
\usepackage{graphicx}
\usepackage{amsmath}
\usepackage{amssymb}

\usepackage{multirow}
\usepackage{caption}
\usepackage{subfigure}

\usepackage[breaklinks=true,bookmarks=false]{hyperref}
\hypersetup{
	colorlinks=true,
	linkcolor=red,
	filecolor=blue,      
	urlcolor=red,
	citecolor=green,
}

\iccvfinalcopy 

\def\iccvPaperID{1370} 

\ificcvfinal\fi

\begin{document}

\title{Temporal Knowledge Propagation for Image-to-Video Person Re-identification}

\author{Xinqian Gu$^{1,2}$, Bingpeng Ma$^{2}$, Hong Chang$^{1,2}$, Shiguang Shan$^{1,2,3}$, Xilin Chen$^{1,2}$\\
$^1$Key Laboratory of Intelligent Information Processing of Chinese Academy of Sciences (CAS),\\Institute of Computing Technology, CAS, Beijing, 100190, China\\
$^2$University of Chinese Academy of Sciences, Beijing, 100049, China\\
$^3$CAS Center for Excellence in Brain Science and Intelligence Technology, Shanghai, 200031, China\\
{\tt\small xinqian.gu@vipl.ict.ac.cn, bpma@ucas.ac.cn, \{changhong, sgshan, xlchen\}@ict.ac.cn}
}

\maketitle
\ificcvfinal\thispagestyle{empty}\fi

\begin{abstract}
In many scenarios of Person Re-identification (Re-ID), the gallery set consists of lots of surveillance videos and the query is just an image, thus Re-ID has to be conducted between image and videos.
Compared with videos, still person images lack temporal information.
Besides, the information asymmetry between image and video features increases the difficulty in matching images and videos.
To solve this problem, we propose a novel Temporal Knowledge Propagation (TKP) method which propagates the temporal knowledge learned by the video representation network to the image representation network.
Specifically, given the input videos, we enforce the image representation network to fit the outputs of video representation network in a shared feature space.
With back propagation, temporal knowledge can be transferred to enhance the image features and the information asymmetry problem can be alleviated.
With additional classification and integrated triplet losses, our model can learn expressive and discriminative image and video features for image-to-video re-identification.
Extensive experiments demonstrate the effectiveness of our method and the overall results on two widely used datasets surpass the state-of-the-art methods by a large margin.
Code is available at: \url{https://github.com/guxinqian/TKP}
\end{abstract}

\section{Introduction}

\begin{figure}
	\centering
	\includegraphics[width = 1\columnwidth]{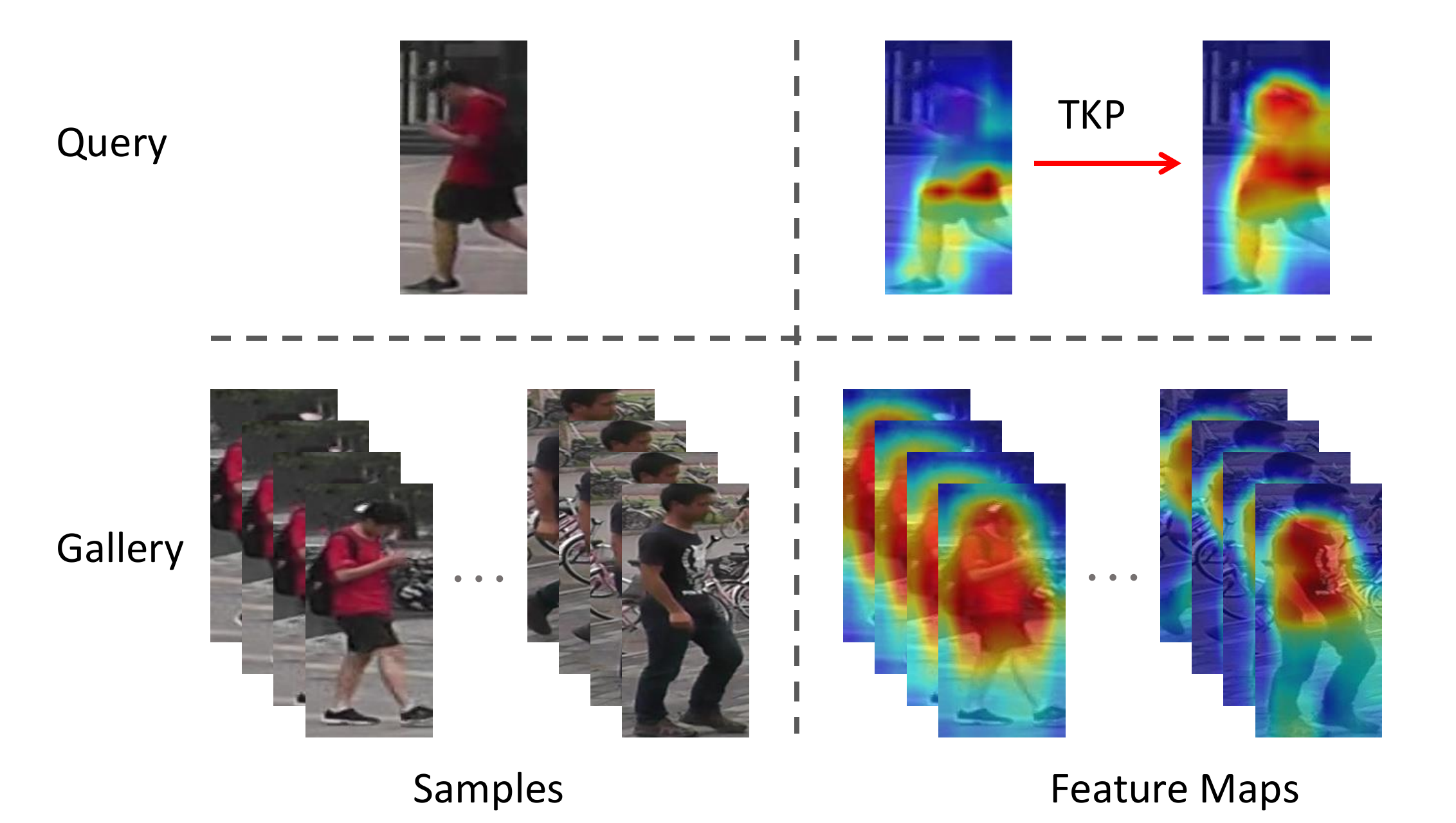}\\
	\vspace{-5pt}
	\captionsetup{font={small}}
	\caption{Visualisation of the feature maps. Due to lack of temporal information, compared with video features, the query image feature only pays attention to some local regions.
	With the proposed TKP method, temporal knowledge can be transferred to enhance the image feature and the learned image feature focuses on more foreground of person image.}
	\vspace{-15pt}
	\label{fig:introduction}
\end{figure}

Person Re-identificaiton (Re-ID) aims to find the sample among the gallery set which has the same identity with a given query.
In general, Re-ID problems fall into two categories: image-based Re-ID~\cite{Bai2017Scalable,Yu2018Hard,Hou2019Interaction,Ge2018Deep} and video-based Re-ID~\cite{Wu2018One,Wang2014Person,Hou2019vrstc}.
The main difference is that the query and gallery in image-based Re-ID are both images, while the query and gallery in video-based Re-ID are both videos.

However, in many real-world scenarios, the gallery set is usually constituted by lots of surveillance videos, while the query only consists of one image, thus Re-ID has to be conducted between image and videos.
One instance is rapidly locating and tracking a criminal suspect among a mass of surveillance videos according to one photo of the suspect (e.g., Boston marathon bombings event).
Due to its crucial role in video surveillance system, image-to-video (I2V) Re-ID~\cite{Zhu2017Learning,Zhu2018Image} has attracted increasing attention in recent years.

In I2V Re-ID, the query is a still image, while the gallery of videos contains additional temporal information.
Some research~\cite{Zhang2017Image,Zhu2017Learning} indicates that modeling temporal relations between video frames makes the gallery video features robust to disturbing conditions.
However, these methods ignore that the query of I2V Re-ID only consists of one still image and lacks temporal information.
As a result, on one hand, the image feature cannot benefit from the advantages of modeling temporal relations (see Figure~\ref{fig:introduction}).
On the other hand, the information asymmetry between image and video features increases the difficulty in measuring the image-to-video similarity.
Hence, it is essential and desirable to develop a method to supplement temporal information to image feature representation.

In this paper, we propose a novel Temporal Knowledge Propagation (TKP) method to address the problems of ignored image temporal representation and information asymmetry in I2V Re-ID.
This is inspired by knowledge distillation~\cite{Hinton2015Distilling}, which transfers dark knowledge from a large and powerful teacher network to a smaller and faster student network.
In our TKP method, the temporal knowledge learned by video representation network is propagated to image representation network.
During training, given the same input videos, we enforce the frame features extracted by the image representation network to match the outputs of video representation network in a shared feature space.
After training with back propagation, the temporal knowledge can be naturally transferred from video representation network to image representation network.
In the test stage, we use the trained image representation network to extract the query image features.
Thanks to the transferred temporal knowledge, the extracted image features manifest robustness to disturbing conditions just like the video frame features (see Figure~\ref{fig:introduction}).
Meanwhile, the information asymmetry problem between image and video features is addressed, thus it is much easier to measure the similarity between images and videos.

Extensive experiments validate the effectiveness of the proposed method.
For instance, on MARS dataset, our method improves the performance from 67.1\% to 75.6\% (+8.5\%) \wrt top-1 accuracy, surpassing the state-of-the-art methods by a large margin.

\section{Related Work}

\noindent
{\bf I2V Re-ID.}
Recently, several relevant methods~\cite{Wang2017P2SNet,Zhang2017Image,Zhu2017Learning,Zhu2018Image} are proposed for I2V Re-ID task.
Among them, Zhu \etal~\cite{Zhu2017Learning,Zhu2018Image} firstly investigate this problem and propose a heterogeneous dictionary pair learning framework to map image features and video features to a shared feature space.
Wang \etal~\cite{Wang2017P2SNet} try to use deep learning based methods to solve this problem and Zhang \etal~\cite{Zhang2017Image} use LSTM to model the temporal information of gallery videos to enhance the robustness of the video features.
However, these methods~\cite{Zhu2017Learning, Zhu2018Image,Zhang2017Image} neglect that there is no temporal information in query image features.
In contrast, our proposed TKP method transfers temporal knowledge learned by video representation network to image representation network, which can effectively reduce the information asymmetry between image and video features.

\vspace{5pt}
\noindent
{\bf Modeling Temporal Relations.}
Dealing with temporal relations between video frames is of central importance in video feature extraction.
A natural solution is to apply Recurrent Neural Networks (RNN) to model sequences~\cite{Chung2017A, Mclaughlin2016Recurrent, Ng2015Beyond}.
There are also some methods~\cite{Feichtenhofer2017Spatiotemporal, Shuiwang20103D, Du2015Learning} that use 3D convolution (C3D) to process temporal neighborhoods.
However, both RNN and C3D only process one local neighborhood at a time.
Recently, Wang \etal~\cite{Wang2018Non} propose to use non-local operation to capture long-range dependencies, which achieves higher results on video classification task.
In this paper, we also attempt to utilize non-local operation to model temporal relations in person videos.

\vspace{5pt}
\noindent
{\bf Knowledge Distillation.}
Knowledge distillation~\cite{Cristian2006Model,Hinton2015Distilling,Romero2015FitNets,Chen2018DarkRank} is a widely used technique to transfer knowledge from a teacher network to a student network.
Generally, it is used to transfer from a powerful and large network to a faster and small network.
In contrast, our TKP method is the first time to transfer temporal knowledge from video representation network to image representation network.
Besides, instead of using a well-trained teacher, our image-to-video representation learning and temporal knowledge transferring are trained simultaneously.

As for the distillation forms, Hinton \etal~\cite{Hinton2015Distilling,Zhang2018Deep} minimize the Kullback-Leibler divergence of the final classification probability of teacher network and student network to transfer knowledge.
In contrast, Bengio \etal~\cite{Romero2015FitNets} directly minimize the Mean Square Error (MSE) of the middle outputs of these two networks.
For deep metric learning tasks, Chen \etal~\cite{Chen2018DarkRank, Zhang2017AlignedReID} transfer knowledge via cross sample similarities.
In this paper, we transfer temporal knowledge by minimizing the MSE of the image features and the corresponding video frame features in a shared feature space, which is  similar to \cite{Romero2015FitNets} in loss design but different in models. Besides, we also formulate the TKP loss based on cross sample distances in the shared feature space.


\begin{figure*}
   \centering
   \includegraphics[width = 1.9\columnwidth]{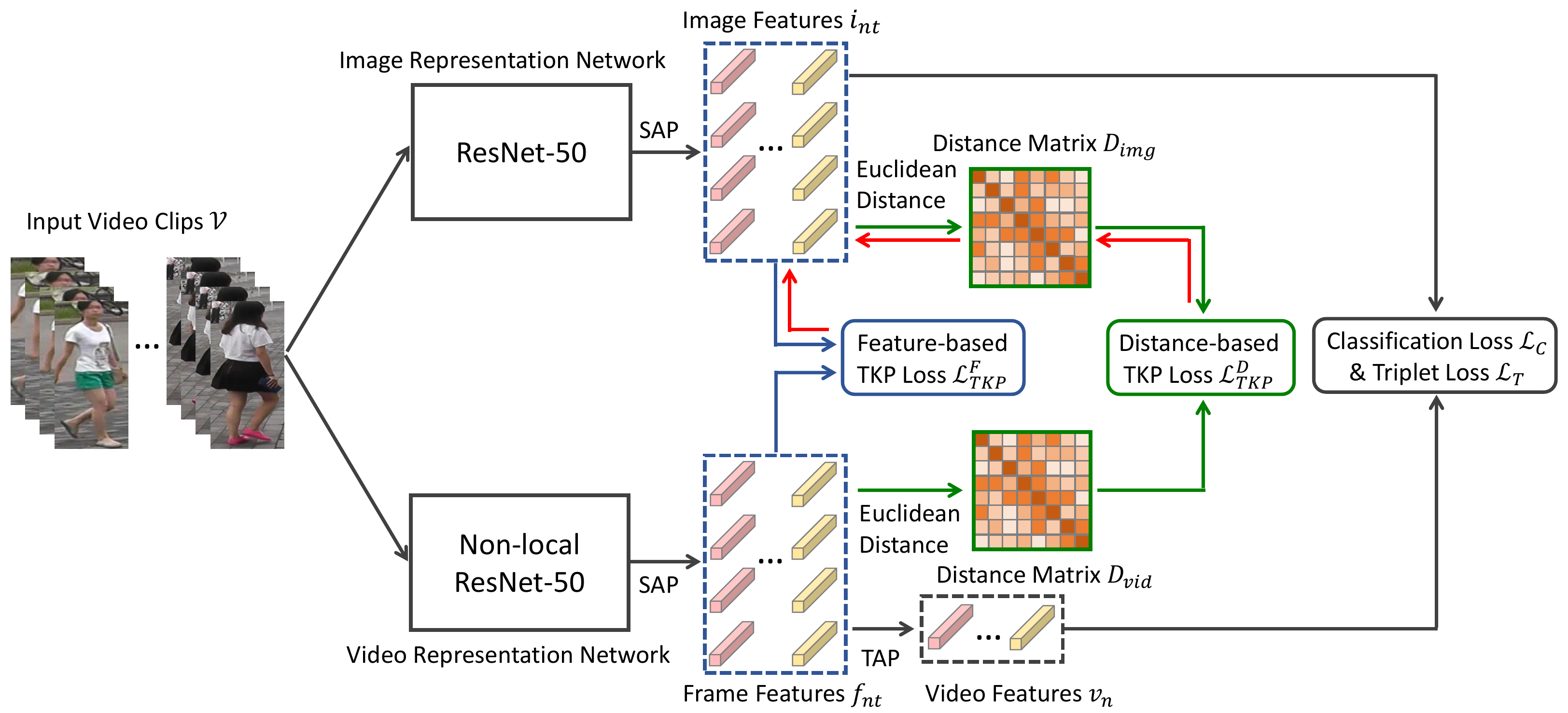}\\
   \vspace{-5pt}
   \captionsetup{font={small}}
   \caption{The framework of TKP method in I2V Re-ID training.
  	SAP and TAP represent spatial average pooling and temporal average pooling respectively.
   The classification loss and triplet loss are used to guide image-to-video representation learning.
   The blue arrow represents the process of TKP via features, while the green arrow represents the process of TKP via cross sample distances.
   And the red arrow denotes the back propagation process of TKP loss.
   The representation learning and temporal knowledge transferring are trained simultaneously.
   Best viewed in color. }
   \vspace{-15pt}
   \label{fig:network_architecture}
\end{figure*}

\section{The Proposed Method}
In this section, we first introduce the overall network architecture for the proposed TKP method.
Secondly, the details of image representation network and video representation network are illustrated.
Then, our TKP method is presented, followed by the final objective function and sampling strategy.
Finally, these two learned networks are used to perform I2V Re-ID testing.

The framework for our proposed TKP method in I2V Re-ID training is depicted in Figure~\ref{fig:network_architecture}.
Given input video clips, the image representation network extracts visual information of single-frame images, while the video representation network extracts visual information and deals with temporal relations between video frames simultaneously.
Temporal knowledge propagation from the video representation network to image representation network is formulated by constructing the TKP loss.
By minimizing the TKP loss, together with the classification and triplet losses, the image features and video features are mapped to a shared feature space.
Details of the proposed method are given as follows.

\subsection{Image Representation Network}
\label{sec:imagemodel}

We use ResNet-50~\cite{He2016Deep} without final fully-connected layer as image representation network for visual feature learning.
To enrich the granularity of image features, we remove the last down-sampling operation of ResNet-50 following \cite{Sun2018Beyond}.

Given $N$ person video clips $\mathcal{V}=\{V_n\}^N_{n=1}$, each $V_n$ contains $T$ frames $V_n=\{F_{n1},F_{n2},...,F_{nT}\}$ (Unless specified, we set $T=4$).
If we discard the temporal relations between video frames, these video clips $\mathcal{V}$ can be considered as a set of individual images $\{F_{nt}\}_{n=1,t=1}^{N,T}$.
Hence, we can use the image representation network $\mathcal{F}_{img}(\cdot)$ to extract features of these images, for all $n,t$,
\begin{equation}\label{eq:imagenetwork}
	i_{nt}=\mathcal{F}_{img}(F_{nt}),
\end{equation}
where $i_{nt}\in \mathbb{R}^{D}$ is the corresponding image feature of video frame $F_{nt}$. As for ResNet-50, $D$ is $2048$.

\subsection{Video Representation Network}
\label{sec:videomodel}

To model the visual and temporal information of video sequences simultaneously, we combine CNN with non-local neural network~\cite{Wang2018Non} as video representation network.
The non-local block computes the response at a position as a weighted sum of the features at all positions in the input feature map.
It can naturally handle the temporal relations between video frames.

\begin{table}
   \small
   \captionsetup{font={small}}
   \caption{The architecture of video representation network. Each input video clip contains 4 frames, each frame with $256\times 128$ pixels.}
   \vspace{-20pt}
   \begin{center}
   \setlength{\tabcolsep}{1.5mm}{
      \begin{tabular}{c | c | c}
         \hline
         \multicolumn{2}{c|}{layer} &output size \\
         \hline
         conv$_1$  &$7\times7$, stride 2, 2     &$4\times128\times64\times64$\\
         \hline
         pool   &$3\times3$ max, stride 2, 2     &$4\times64\times32\times64$ \\
         \hline
         res$_2$&residual block $\times3$           &$4\times64\times32\times256$\\
         \hline
         res$_3$ &$\left[\begin{array}{ll}   \text{residual block} \times 2 \\   \text{non-local block} \times 1 \\
         \end{array}\right]\times2$  &$4\times32\times16\times512$\\[1.5 ex]
         \hline
         res$_4$ &$\left[\begin{array}{ll}   \text{residual block} \times 2 \\   \text{non-local block} \times 1 \\
         \end{array}\right]\times3$  &$4\times16\times8\times1024$\\[1.5 ex]
         \hline
         res$_5$ &residual block $\times3$  &$4\times16\times8\times2048$\\
         \hline
         \multicolumn{2}{c|}{spatial average pool}  &$4\times2048$ \\
         \hline
         \multicolumn{2}{c|}{temporal average pool}  &$2048$\\
         \hline
      \end{tabular}}
   \end{center}
   \vspace{-15pt}
   \label{tab:vidmodel}
\end{table}

Table~\ref{tab:vidmodel} shows the model structure of our video representation network on the backbone of ResNet-50.
Specifically, we add two non-local blocks to $\text{res}_3$ and three non-local blocks to $\text{res}_4$ and remove the last down-sampling operation in $\text{res}_5$ to enrich granularity.
Given an input video clip $V_n=\{F_{n1},F_{n2},...,F_{nT}\}$ with $T$ frames, the video representation network $\mathcal{F}_{vid}(\cdot)$ is defined as:
\begin{equation}\label{eq:videonetwork}
	\{f_{n1},f_{n2},...,f_{nT}\}=\mathcal{F}_{vid}(F_{n1},F_{n2},...F_{nT}),
\end{equation}
where $f_{nt} \in \mathbb{R}^{D}, t=1,...,T$ is the video frame feature of $F_{nt}$.
With temporal average pooling, multiple video frame features of a video clip can be integrated to a video feature $v_n\in \mathbb{R}^{D}$.

\subsection{Temporal Knowledge Propagation}
\label{sec:tkp}

In general, the performance of Re-ID highly depends on the robustness of feature representation.
It has been proved that modeling temporal relations between video frames makes the person appearance representations robust to large variations~\cite{You2016Top}.
However, image representation network takes in still images and cannot process temporal relations, thus the output image features cannot benefit from temporal knowledge.
To solve this problem, we propose TKP method which enforces the outputs of image representation network to fit the robust outputs of video representation network in the shared feature space.
By back propagation algorithm, the image representation network can learn temporal knowledge from the video frame features.
Consequently, the features extracted by image representation network are assigned, though not directly, some video temporal information.

Specifically, given the input video clips $\mathcal{V}$, we can use Eq.~\eqref{eq:imagenetwork} and Eq.~\eqref{eq:videonetwork} to extract image features $i_{nt}$ and video frame features $f_{nt}$ for all $n=1,...,N, t=1,...,T$.
Since $\mathcal{F}_{vid}(\cdot)$ extracts visual information and deals with temporal relations between video frames simultaneously, $f_{nt}$ not only contains the visual information of video frame $F_{nt}$, but also involves temporal relations with other frames.
In order to use video frame feature $f_{nt}$ to propagate the temporal knowledge to image representation network,
we formulate the TKP method as an optimization problem from the following two ways.

\vspace{5pt}
\noindent
{\bf Propagation via Features.} The first way is enforcing image representation network to fit the robust video frame features in a shared feature space. In this case, TKP method can be formulated to minimize the MSE between the image features and the corresponding video frame features:
\begin{equation}\label{eq:tkpviafeature}
	\mathcal{L}^F_{TKP}=\frac{1}{N T} \sum_{n=1}^{N} \sum_{t=1}^{T}\| i_{nt}-f_{nt} \|_2^2,
\end{equation}
where $\| \cdot \|_2$ denotes $l_2$ distance.

Eq.~\eqref{eq:tkpviafeature} can be considered as simplified \emph{moving least squares}~\cite{Levin1998The,Schaefer2006Image}, which is capable of reconstructing continuous functions from a set of labeled samples. Here, our target is to reconstruct the image representation function $\mathcal{F}_{img}(\cdot)$ from the video frame representations ${(F_{nt},f_{nt})}_{n=1,t=1}^{N,T}$.
This formulation is similar to FitNets~\cite{Romero2015FitNets}, except that the outputs of teacher network and student network in \cite{Romero2015FitNets} are mapped to the same dimension via an extra convolutional regressor.
In contrast, the outputs of the image and video representation networks in our framework have the same dimension and the network construction is similar, thus we do not need additional convolutional regressor.

\vspace{5pt}
\noindent
{\bf Propagation via Cross Sample Distances.} 
Another way to propagate temporal knowledge from video representations to image representations may resort to neural network embedding.
The structure of the target embedding space is characterized by cross sample distances.
For all video frame features $\{f_{nt}\}_{n=1,t=1}^{N,T}$, we compute the cross sample Euclidean distances matrix as $D_{vid}\in \mathbb{R}^{NT\times NT}$. To estimate the image representation in the embedding space, we constrain that the cross image distances $D_{img}\in \mathbb{R}^{NT\times NT}$ are consistent with the cross video frame distances  $D_{vid}$. In this way, the temporal information, as well as the sample distribution, are propagated to the image representation network. The TKP loss is formulated as:
\begin{equation}\label{eq:tkpviacsd}
	\mathcal{L}^D_{TKP}=\frac{1}{N T} \| D_{img}- D_{vid} \|_F^2,
\end{equation}
where $\| \cdot \|_F$ denotes Frobenius norm.
This formulation is similar to \emph{multidimensional scaling}~\cite{Kruskal1964Multidimensional}, except that we use a deep network to model the embedding function $\mathcal{F}_{img}(\cdot)$ instead of directly computing the embedded features via eigen-decomposition.

Eq.~\eqref{eq:tkpviafeature} and Eq.~\eqref{eq:tkpviacsd} transfer knowledge from different levels and are complementary to each other.
The empirical comparison of the two ways is provided in Section~\ref{sec:resultsonI2VRe-ID}.

{\bf Note} that both image and video networks use ResNet-50 as backbone.
The only difference is that video network add extra non-local blocks to model temporal information.
Given the same inputs, TKP loss enforces these two networks to output similar features.
Obviously, the weights of additional non-local blocks being 0 is the optimal solution of minimizing TKP loss.
In that case, the non-local blocks can not capture any temporal information.
So updating video network by TKP deteriorates modeling temporal knowledge.
Unless specified, in our emperiments, $\mathcal{L}^F_{TKP}$ and $\mathcal{L}^F_{TKP}$ are not back-propagated through the video representation network during model training.

\subsection{Objective Function}
\label{sec:loss}
Besides the TKP loss, additional identification losses are also needed to learn discriminative features for image-to-video re-identification.
In this paper, we utilize the widely used classification loss and integrated triplet loss.
In fact, other identification losses are also applicable.

\vspace{3pt}
\noindent
{\bf Classification Loss.} Considering person identities as category-level annotations, we build two shared weights classifiers to map the image features and video features to a shared identity space.
The classifiers are implemented as a linear layer following by a softmax operation and the output channel is the number of identities of training set.
The classification loss $\mathcal{L}_{C}$ can be formulated as the cross entropy error between the predicted identities and the correct labels.

\vspace{3pt}
\noindent
{\bf Integrated Triplet Loss.} We also use triplet loss with hard sample mining~\cite{Hermans2017In} to constrain the relative sample distances in the shared feature space.
Specifically, we integrate four kinds of triplet losses, image-to-video (I2V), video-to-image (V2I), image-to-image (I2I) and video-to-video (V2V) triplet losses.
The final triplet loss $\mathcal{L}_{T}$ is defined as:
\begin{equation}\label{eq:tripletloss}
\mathcal{L}_{T}= \mathcal{L}_{I2V}+\mathcal{L}_{V2I}+\mathcal{L}_{I2I}+\mathcal{L}_{V2V},
\end{equation}
where
\begin{equation}\label{eq:I2Vloss}
\mathcal{L}_{I2V}=\Big[ m + \max_{v_p\in S_a^+} d(i_a, v_p) - \min_{v_n\in S_a^-} d(i_a, v_n) \Big]_+,
\end{equation}
\begin{equation}\label{eq:V2Iloss}
\mathcal{L}_{V2I}=\Big[ m + \max_{i_p\in S_a^+} d(v_a, i_p) - \min_{i_n\in S_a^-} d(v_a, i_n) \Big]_+,
\end{equation}
\begin{equation}\label{eq:I2Iloss}
\mathcal{L}_{I2I}= \Big[ m + \max_{i_p\in S_a^+} d(i_a, i_p) - \min_{i_n\in S_a^-} d(i_a, i_n) \Big]_+,
\end{equation}
\begin{equation}\label{eq:V2Vloss}
\mathcal{L}_{V2V}= \Big[ m + \max_{v_p\in S_a^+} d(v_a, v_p) - \min_{v_n\in S_a^-} d(v_a, v_n) \Big]_+ .
\end{equation}
Here $m$ is a pre-defined margin, $d(\cdot,\cdot)$ denotes the Euclidean distance, and $[\cdot]_+=\max(0,\cdot)$.
$S_a^+$ and $S_a^-$ are the positive and negative sample sets of the anchor sample ($i_a$ or $v_a$) respectively.

Among the four losses, Eq.~\eqref{eq:I2Vloss} and Eq.~\eqref{eq:V2Iloss} constrain the distance between image feature and video feature, which can improve the between-modality feature discriminativity.
In contrast, Eq.~\eqref{eq:I2Iloss} and Eq.~\eqref{eq:V2Vloss} constrain within-modality relative distances, which makes our model distinguish the fine-grained differences between different identities within the same modality.
The between-modality and within-modality losses are complementary and their integration can improve image-to-video representation learning.

\vspace{5pt}
\noindent
{\bf Objective Function.}
The image-to-video representation learning and temporal knowledge transferring are trained simultaneously.
The final objective function is formulated as the combination of classification loss, integrated triplet loss and the proposed TKP loss:
\begin{equation}\label{eq:loss}
\mathcal{L}=\mathcal{L}_{C} + \mathcal{L}_{T} + \mathcal{L}_{TKP}^F + \mathcal{L}_{TKP}^D.
\end{equation}

\subsection{Sampling Strategy}
\label{sec:sampling}
To better train the model with multiple losses, we design a particular sampling strategy.
For each batch, we randomly select $P$ persons.
For each person, we randomly select $K$ video clips, each with $T$ frames.
All the $P\times K=N$ video clips are fed into video representation network.
Meanwhile, all $N\times T$ frames form an image batch and are fed into image representation network.
In this way, all the samples in a mini batch can be reused to compute these three losses in Eq.~\eqref{eq:loss}, which can reduce the computational cost.

\subsection{Image-to-video Re-ID Testing}
\label{sec:i2vtesting}
\begin{figure}[t]
	\begin{center}
		\includegraphics[width = 0.9\columnwidth]{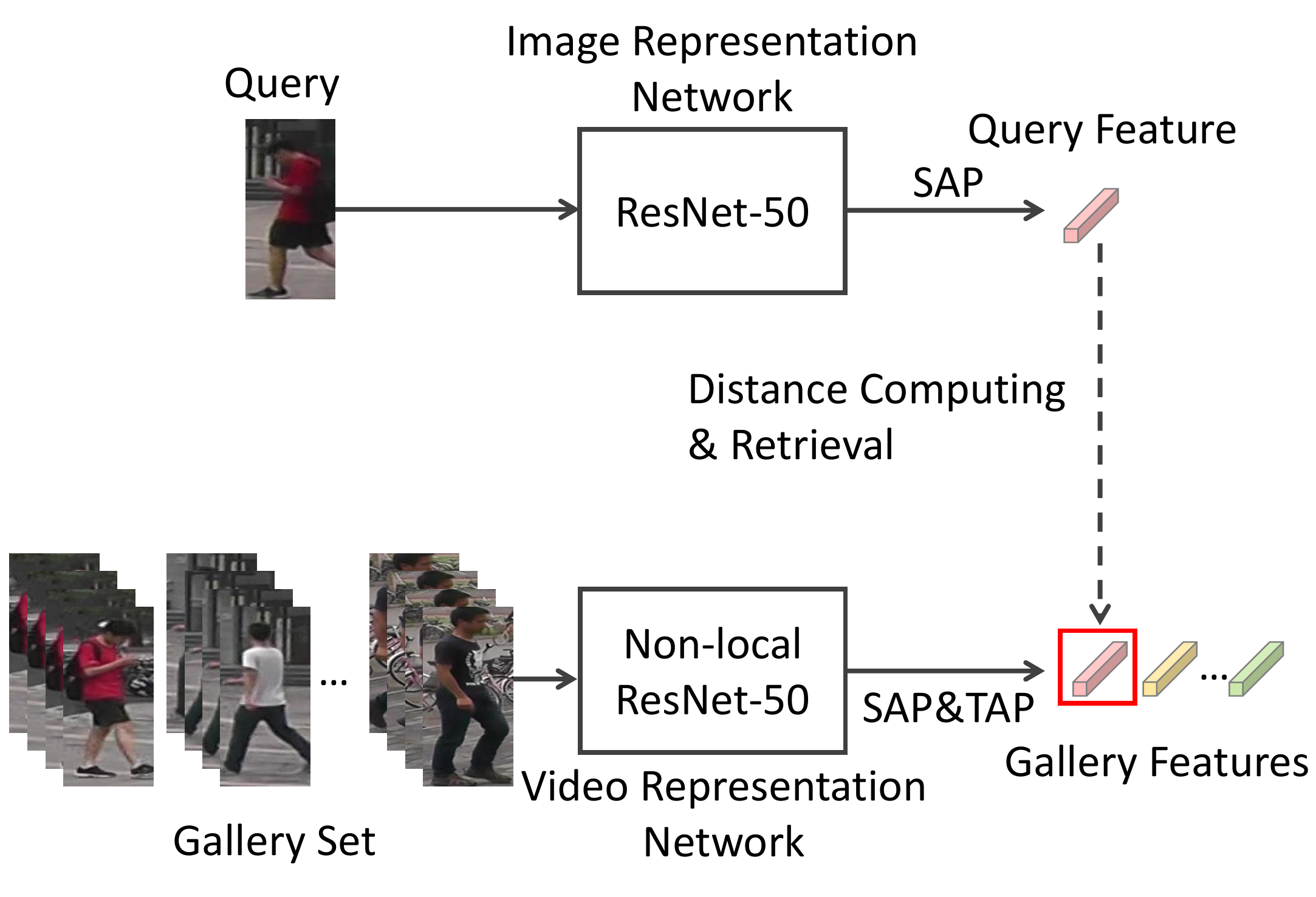}
	\end{center}
	\vspace{-20pt}
	\captionsetup{font={small}}
	\caption{The pipeline of I2V Re-ID testing.}
	\vspace{-20pt}
	\label{fig:I2Vtesting}
\end{figure}

In the test stage, each query is a still image and the gallery set consists of masses of person videos.
The process of image-to-video Re-ID testing is shown in Figure~\ref{fig:I2Vtesting}.
Specifically, we use the learned image representation network after TKP to extract image feature of the query and the gallery video features are extracted by video representation network.
After feature extraction, we compute the distances between the query feature and each gallery video feature and then conduct image-to-video retrieval according to the distances.

\section{Experiments}

\subsection{Datasets and Evaluation Protocol}

\noindent
{\bf Datasets.}
We evaluate our method on MARS~\cite{Zheng2016MARS}, DukeMTMC-VideoReID (Duke)~\cite{Wu2018One} and iLIDS-VID~\cite{Wang2014Person} datasets.
Among them, MARS and Duke are multi-camera datasets, while iLIDS-VID is captured by only two cameras.
The amounts of person videos on MARS, Duke and iLIDS-VID are 20478, 5534 and 600 respectively, and the average lengths of person videos on these three datasets are 58, 168 and 71 respectively.

\vspace{5pt}
\noindent
{\bf Evaluation Protocol.} All the three datasets above are video Re-ID datasets.
For multi-camera datasets (MARS and Duke),  we just use the first frame of every query video as query image to perform I2V Re-ID testing following \cite{Wang2017P2SNet}.
For iLIDS-VID, we use the first frames of all person videos captured by the first camera for both training and testing in order to be consistent with \cite{Zhu2017Learning,Wang2017P2SNet,Zhang2017Image}. 

\begin{table*}
	\centering
	\captionsetup{font={small}}
	\caption{The results of I2V, I2I and V2V Re-ID on the MARS and Duke datasets. In I2I setting, only the first frames of the query and gallery samples are used. In V2V setting, the full-length query videos and gallery videos are used. All the image features in I2I and I2V Re-ID are extracted by the learned image representation network.  All the video features in I2V and V2V Re-ID are extracted by the learned video representation network.}
	\vspace{-20pt}
	\small
	\begin{center}
		\setlength{\tabcolsep}{1.6mm}{
			\begin{tabular}{l |c c c c|c c| c c| c c|c c |c c| c c}
				\hline 
				\multirow{3}*{Models} &\multicolumn{4}{c}{Losses} &\multicolumn{6}{|c|}{MARS}  &\multicolumn{6}{|c}{Duke} \\   
				\cline{2-17}
				&\multirow{2}*{$\mathcal{L}_{C}$} &\multirow{2}*{$\mathcal{L}_{T}$} &\multirow{2}*{$\mathcal{L}_{TKP}^F$} &\multirow{2}*{$\mathcal{L}_{TKP}^D$} &\multicolumn{2}{|c|}{I2V Re-ID} &\multicolumn{2}{|c|}{I2I Re-ID} &\multicolumn{2}{|c|}{V2V Re-ID} &\multicolumn{2}{|c|}{I2V Re-ID} &\multicolumn{2}{|c|}{I2I Re-ID} &\multicolumn{2}{|c}{V2V Re-ID} \\
				\cline{6-17}
				&&&& &top-1  &mAP &top-1  &mAP &top-1  &mAP &top-1  &mAP &top-1  &mAP &top-1  &mAP\\ 
				\hline
				baseline  &\checkmark &\checkmark & & &67.1 &55.5 &65.9 &49.2 &83.4 &72.6 &67.5 &65.6 &60.4 &52.8 &93.2 &91.3\\
				TKP-F     &\checkmark &\checkmark &\checkmark & &75.0 &64.2 &\bfseries71.0 &54.7 &83.2 &72.6 &76.8 &74.2 &63.0 &54.5 &93.6 &91.5\\
				TKP-D     &\checkmark &\checkmark & &\checkmark &75.0 &63.1 &70.3 &\bfseries55.0 &\bfseries84.1 &72.9 &76.5 &74.9 &62.0 &53.5 &93.3 &91.4\\
				TKP       &\checkmark &\checkmark &\checkmark &\checkmark &\bfseries75.6 &\bfseries65.1 &\bfseries71.0 &\bfseries55.0 &84.0 &\bfseries73.3 &\bfseries77.9 &\bfseries75.9 &\bfseries63.4 &\bfseries54.8 &\bfseries94.0 &\bfseries91.7\\
				\hline
		\end{tabular}}
	\end{center}
	\vspace{-20pt}
	\label{tab:comparisonexperiments}
\end{table*}

We use the Cumulative Matching Characteristics (CMC) to evaluate the performance of each approach. 
For iLIDS-VID, the experiment is repeated 10 times and the average result is presented. 
For multi-camera datasets, we also report the mean Average Precision (mAP)~\cite{Zheng2015Scalable} as a complement to CMC.

The comparison experiments are mainly conducted on MARS and Duke, since these two datasets have fixed training/testing splits, which is convenient for extensive evaluation.
We also present the final results on iLIDS-VID as well as MARS to compare with the state-of-the-art methods.

\subsection{Implementation Details}

We pre-train ResNet-50 on ImageNet~\cite{Olga2015ImageNet} and adopt the method in \cite{Wang2018Non} to initialize the non-local blocks.
During training, we randomly sample 4 frames with a stride of 8 frames from the original full-length video to form an input video clip.
For the original video less than 32 frames, we duplicate it to meet the length.
The parameters $P$ and $K$ in Section~\ref{sec:sampling} are both set to 4.
The input video frames are resized to $256\times 128$ pixels.
Only horizontal flip is used for data augmentation. 
We adopt Adaptive Moment Estimation (Adam)~\cite{Kingma2014Adam} with weight decay 0.0005 to optimize the parameters. 
The model is trained for 150 epochs in total.
The learning rate is initialized to 0.0003 and divided by 10 after every 60 epochs.
For iLIDS-VID, we first pre-train the model on large-scale dataset and then fine-tune it on iLIDS-VID following \cite{Wang2017P2SNet}.

In the test phase, the query image features are extracted by image representation model. 
For each gallery video, we first split it into several 32-frame clips. 
For each clip, we utilize video representation model to extract video representation. 
The final video feature is the averaged representation of all clips.

\subsection{Results on I2V Re-ID}
\label{sec:resultsonI2VRe-ID}

To validate the effectiveness of the proposed TKP method for I2V Re-ID, we implement and test a baseline and several variants of our model. 
The configurations of these approaches are presented in Table \ref{tab:comparisonexperiments}.
Among them, \textit{baseline} only adopts classification loss and triplet loss for image-to-video representation learning.
\textit{TKP-F} and \textit{TKP-D} extra use Eq.~\eqref{eq:tkpviafeature} and Eq.~\eqref{eq:tkpviacsd} to transfer temporal knowledge respectively.
\textit{TKP} combines these two transfer ways during training.
The results of I2V Re-ID on the MARS and Duke dataset can be seen in Table~\ref{tab:comparisonexperiments}.

Compared with \textit{baseline}, \textit{TKP-F} and \textit{TKP-D} consistently improve the performance by a large margin. 
Specifically, \textit{TKP-F} increases the mAP by 8.7\% and 8.6\% on MARS and Duke respectively.
For \textit{TKP-D}, the improvement of mAP is 7.6\% on MARS, and 9.3\% on Duke.
This comparison shows that temporal knowledge transfer is essential for image-to-video representation learning.
We also compare the combination method \textit{TKP} with \textit{TKP-F} and \textit{TKP-D}.
It can be seen that a farther improvement is achieved.
This result demonstrates that these two transfer ways transfer temporal knowledge from different perspectives and are complementary to each other.

\subsection{How does TKP Work?}

To investigate how TKP works, we extra use the image representation networks trained in Section~\ref{sec:resultsonI2VRe-ID} to conduct image-to-image (I2I) Re-ID tests, where only the first frames of the original query and gallery videos are utilized.
Moreover, we also use the trained video representation networks to perform video-to-video (V2V) Re-ID experiments, where the original full-length query and gallery videos are used. 
The experiment results are also presented in Table~\ref{tab:comparisonexperiments}.

Compared with \textit{baseline}, different transfer methods consistently improve the I2I Re-ID performance on both MARS and Duke dataset.
Especially, \textit{TKP} increases the mAP from 49.2\% to 55.0\% (+5.8\%) on the  MARS dataset.
Moreover, the V2V Re-ID performance of different transfer methods is close to \textit{baseline}.
The comparisons demonstrate that the proposed TKP method can improve the robustness of learned image features, meanwhile, does not reduce the discriminativity of video features.
In addition, with the transferred temporal knowledge, the information asymmetry between image and video features can also be alleviated, thus the I2V Re-ID performance gains more improvement.

\subsection{Comparison among I2I, I2V and V2V Re-ID}

\begin{figure}[t]
	\begin{center}
		\includegraphics[width = 1\columnwidth]{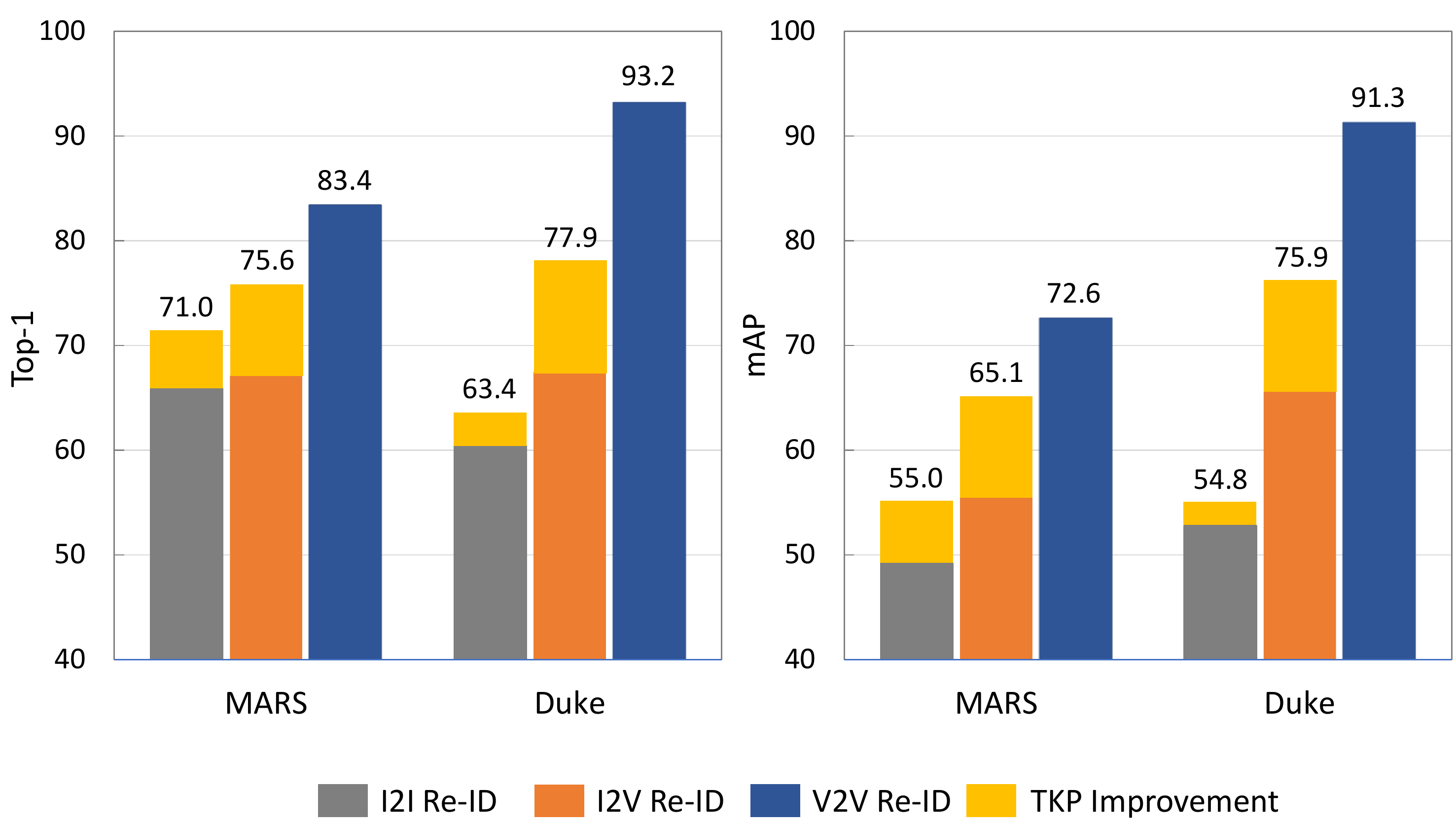}
	\end{center}
	\vspace{-15pt}
	\captionsetup{font={small}}
	\caption{Comparison among I2I, I2V and V2V Re-ID on the MARS and Duke datasets. With the proposed TKP method, the performance gap can be significantly reduced.}
	\vspace{-15pt}
	\label{fig:comparei2ii2vv2v}
\end{figure}

\begin{table}
	\captionsetup{font={small}}
	\caption{Comparison with state-of-the-art I2V Re-ID methods on the iLIDS-VID dataset.}
	\vspace{-20pt}
	\small
	\begin{center}
		\setlength{\tabcolsep}{1.5mm}{
			\begin{tabular}{l |c c c c}
				\hline 
				Models	&top-1 &top-5 &top-10 &top-20  \\ 
				\hline
				PSDML~\cite{Zhu2013From}  &13.5 &33.8 &45.6 &56.3 \\
				LERM~\cite{Huang2014Learning}   &15.3 &37.1 &49.7 &62.0 \\
				XQDA~\cite{Liao2015Person}   &16.8 &38.6 &52.3 &63.6 \\
				KISSME~\cite{K2012Large} &17.6 &41.7 &55.3 &68.7 \\
				PHDL~\cite{Zhu2017Learning}          &28.2 &50.4 &65.9 &80.4 \\
				\hline
				TMSL~\cite{Zhang2017Image}       &39.5  &66.9 &79.6 &86.6 \\
				P2SNet~\cite{Wang2017P2SNet}          &40.0  &68.5 &78.1 &90.0 \\
				\hline
				TKP      &\bfseries54.6  &\bfseries79.4 &\bfseries86.9 &\bfseries93.5 \\
				\hline
		\end{tabular}}
	\end{center}
	\vspace{-15pt}
	\label{tab:sotaoniLIDS}
\end{table}

\begin{table}
	\captionsetup{font={small}}
	\caption{Comparison with state-of-the-art I2V Re-ID methods on the MARS dataset.}
	\vspace{-20pt}
	\small
	\begin{center}
		\setlength{\tabcolsep}{1.5mm}{
			\begin{tabular}{l |c c c c}
				\hline 
				Models  &top-1 &top-5 &top-10 &mAP \\ 
				\hline
				P2SNet~\cite{Wang2017P2SNet}     &55.3  &72.9 &78.7 &-\\
				ResNet-50~\cite{He2016Deep}+XQDA~\cite{Liao2015Person}   &67.2 &81.9 &86.1 &54.9\\
				\hline
				TKP      &\bfseries75.6  &\bfseries87.6 &\bfseries90.9 &\bfseries65.1\\
				\hline
		\end{tabular}}
	\end{center}
	\vspace{-15pt}
	\label{tab:sotaonMARS}
\end{table}

\begin{table}
	\captionsetup{font={small}}
	\caption{Comparison with state-of-the-art V2V Re-ID methods on the MARS dataset.}
	\vspace{-20pt}
	\small
	\begin{center}
		\setlength{\tabcolsep}{1.5mm}{
			\begin{tabular}{l |c c c c}
				\hline 
				Models  &top-1 &top-5 &top-10 &mAP \\ 
				\hline
				SDM~\cite{Zhang2018Multi}    &71.2  &85.7 &91.8 &-\\
				MGCAM~\cite{Song2018Mask}    &77.2  &- &- &71.2\\
				DuATM~\cite{Si2018Dual}    &78.7  &90.9 &- &62.3\\
				multi-snippets~\cite{Chen2018Video}  &81.2 &92.1 &- &69.4\\
				DRSA~\cite{Li2018Diversity}    &82.3  &- &- &65.8\\
				\hline
				TKP      &\bfseries84.0  &\bfseries93.7 &\bfseries95.7 &\bfseries73.3\\
				\hline
		\end{tabular}}
	\end{center}
	\vspace{-20pt}
	\label{tab:sotav2vonMARS}
\end{table}

I2I (image-based) Re-ID is a task where the query and each gallery are both images, while the query and each gallery in V2V (video-based) Re-ID are both videos. In I2V setting, the query is an image while each gallery is a video.
We compare the three different tasks under the same configurations and the comparisons on the MARS and Duke dataset are shown in Figure~\ref{fig:comparei2ii2vv2v}.
Due to the lack of additional visual and temporal information, the performance of I2V Re-ID is lower than that of V2V Re-ID, and I2I Re-ID is lower than I2V Re-ID.
Especially, the performance gap on Duke is much larger.
The reason can be attributed that the average length of the videos on Duke is longer than that on MARS.
When we only use one frame of the original video to conduct I2I and I2V tests, the information loss is more serious.
But our proposed TKP method can transfer temporal knowledge to image features, thus the performance gap can be greatly reduced on these two datasets.

\subsection{Comparison with State-of-the-art Methods}

We compare the proposed approach with state-of-the-art I2V Re-ID methods on the iLIDS-VID and MARS datasets.
The results are presented in Table~\ref{tab:sotaoniLIDS} and Table~\ref{tab:sotaonMARS} respectively.
Among them, \textit{PSDML}~\cite{Zhu2013From}, \textit{LERM}~\cite{Huang2014Learning}, \textit{XQDA}~\cite{Liao2015Person}, \textit{KISSME}~\cite{K2012Large} and \textit{PHDL}~\cite{Zhu2017Learning} are handcrafted feature based methods, while \textit{ResNet-50}~\cite{He2016Deep}\textit{+XQDA}~\cite{Liao2015Person}, \textit{TMSL}~\cite{Zhang2017Image} and \textit{P2SNet}~\cite{Wang2017P2SNet} are deep learning based methods.
It can be seen that deep learning based methods significantly outperform the traditional methods with handcrafed features, while our method further surpasses the existing deep learning based methods by a large margin.
Since Duke is a newly released dataset, existing methods have not conducted I2V Re-ID experiments on it. 
Therefore, we do not compare with state-of-the-art methods on this dataset.
Anyway, the results of our method can be seen in Table~\ref{tab:comparisonexperiments}.

Note that the V2V Re-ID performance decides the upper bound of the I2V Re-ID performance.
We also compare the proposed approach with state-of-the-art V2V Re-ID methods on the MARS dataset.
To compare fairly, for \textit{multi-snippets}~\cite{Chen2018Video}, we use the results without optical flow.
As shown in Table~\ref{tab:sotav2vonMARS}, our TKP consistently outperforms these methods.
As for the iLIDS-VID dataset, since we do not use all of the training set (only use the first frames of all videos captured by the first camera), we do not compare V2V Re-ID results with these methods on this dataset.

\begin{figure}[t]
	\centering
	\includegraphics[width = 0.8\columnwidth]{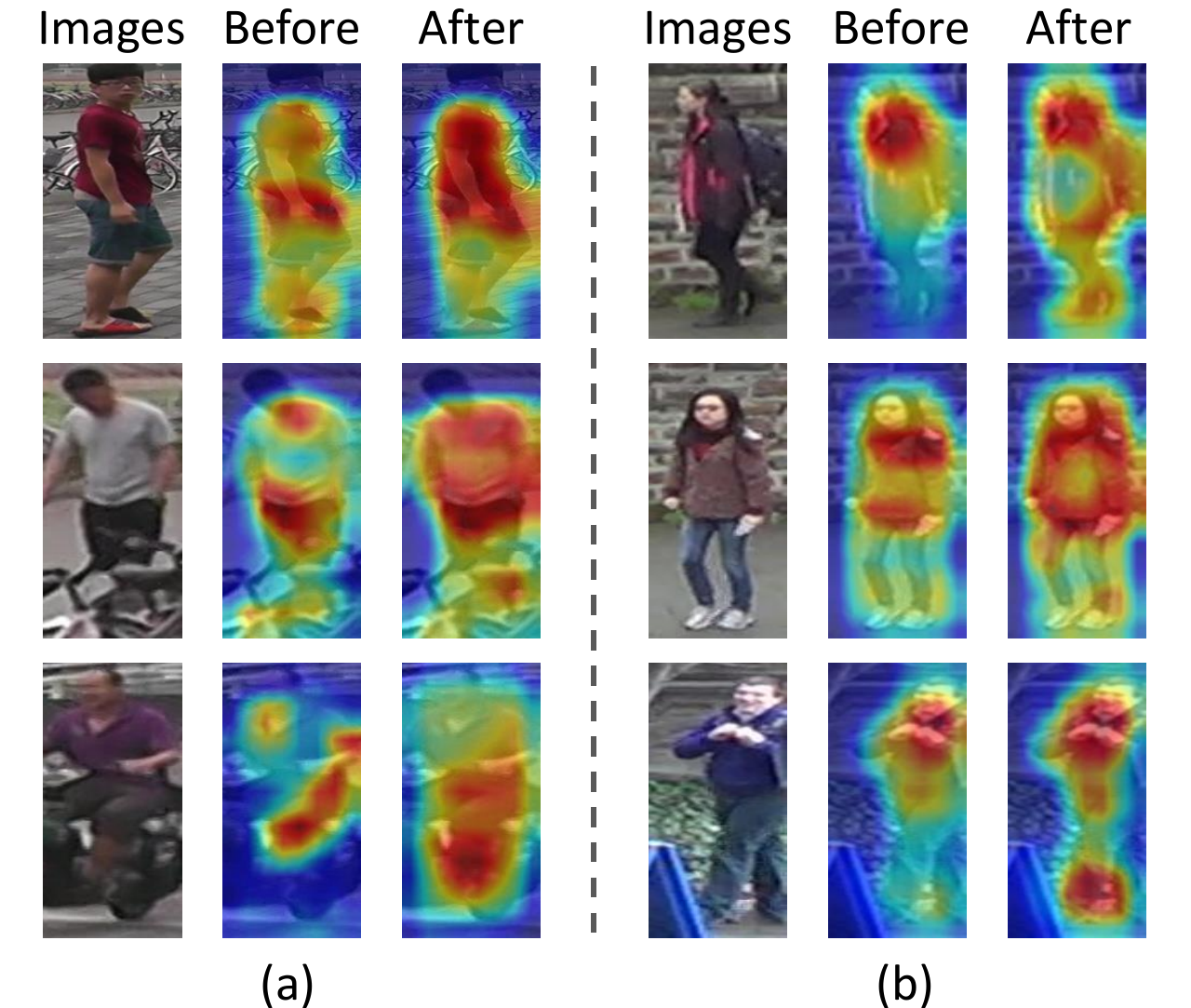}\\
	\vspace{-10pt}
	\captionsetup{font={small}}
	\caption{The visualisation of feature maps before/after TKP on the (a) MARS and (b) Duke datasets. Best viewed in color.}
	\vspace{-15pt}
	\label{fig:fituremap}
\end{figure}

\begin{figure}[t]
	\centering
	\includegraphics[width = 0.9\columnwidth]{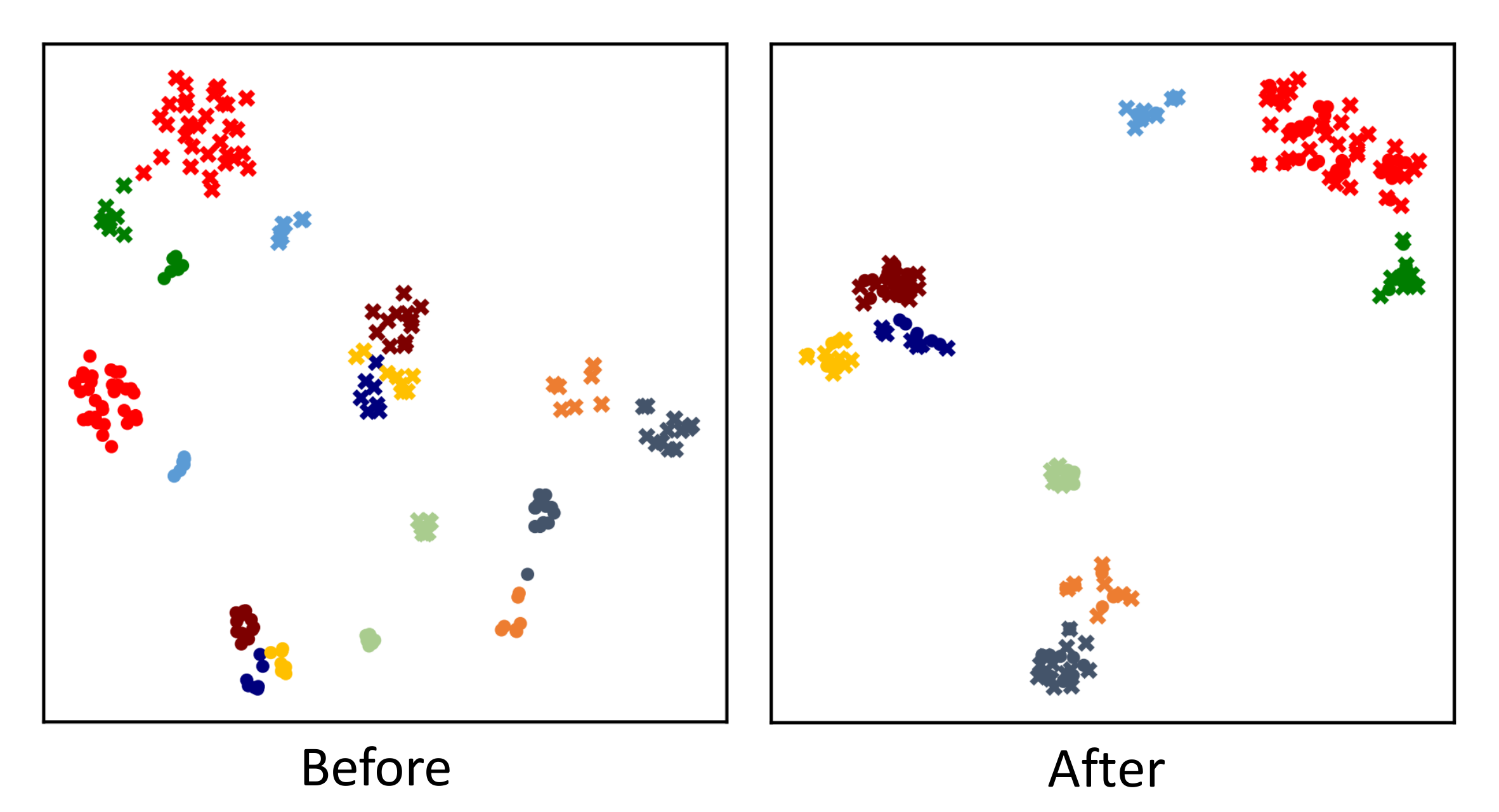}
	\vspace{-10pt}
	\captionsetup{font={small}}
	\caption{The visualisation of feature distribution before/after TKP on the MARS dataset. 
		Circle represents video feature, while cross represents image feature.
		Different colors denote different identities.
		Best viewed in color.}
	\vspace{-15pt}
	\label{fig:featuredistribution}
\end{figure}

\subsection{Visualization}

\noindent
{\bf The visualization of feature maps.} 
We visualise the feature maps of the image features before/after TKP in Figure~\ref{fig:fituremap}.
It can be seen that the original image features only pay attention to some local discriminative regions.
After TKP transferring temporal knowledge, the learned image representations can focus on more foreground and manifest rubustness to occlusion and blur just like video features in Figure~\ref{fig:introduction}, which benefits I2V matching. 
So the final I2V performance can be improved significantly.

\vspace{5pt}
\noindent
{\bf The visualization of feature distribution.} 
We also visualize the distribution of the learned features before/after TKP using t-SNE~\cite{Maaten2008Visualizing} as shown in Figure~\ref{fig:featuredistribution}. 
Before temporal knowledge transferring, the image features and video features with the same identity are incompact.
After TKP transferring temporal knowledge, the feature distributions of these two modalities become more consistent. 
Therefore, it is easier to measure the similarity between image and video features.

\subsection{Ablation Study}
\label{sec:ablationstudy}

\begin{table}[t]
	\centering
	\small
	\captionsetup{font={small}}
	\caption{The I2V Re-ID results with/without non-local blocks on the MARS dataset. w/ NL? denotes whether the model contains non-local blocks or not. The performance improvement is provided in brackets.}
	\vspace{-20pt}
	\begin{center}
		\setlength{\tabcolsep}{3mm}{
			\begin{tabular}{l |c|l l}
				\hline
				Models &w/ NL? &top-1  &mAP\\   
				\hline				
				baseline &-             &66.1 &51.8   \\
				TKP-F &- &68.9(+2.8) &57.8(+6.0)\\
				\hline
				baseline &\checkmark    &67.1 &55.5   \\
				TKP-F &\checkmark &\bfseries75.0(+7.9) &\bfseries64.2(+8.7)\\
				\hline
		\end{tabular}}
	\end{center}
	\vspace{-15pt}
	\label{tab:nonlocal}
\end{table}

\begin{table}[t]
	\centering
	\small
	\captionsetup{font={small}}
	\caption{The results with/without the TKP loss propagating gradient to video representation network on the MARS dataset. BP2v? denotes whether the gradient of TKP loss is propagated to video representation network or not.}
	\vspace{-20pt}
	\begin{center}
		\setlength{\tabcolsep}{1.2mm}{
			\begin{tabular}{l|c |c c|c c|c c}
				\hline
				\multirow{2}*{Models} &\multirow{2}*{BP2v?} &\multicolumn{2}{|c|}{I2I Re-ID}  &\multicolumn{2}{|c}{I2V Re-ID} &\multicolumn{2}{|c}{V2V Re-ID}\\   
				\cline{3-8}				
				& &top-1  &mAP &top-1  &mAP &top-1  &mAP \\
				\hline
				baseline &             &65.9 &49.2 &67.1 &55.5 &83.4 &72.6 \\
				\hline
				\multirow{2}*{TKP-F}&\checkmark    &66.6 &51.0 &72.7 &60.3 &78.5 &66.6 \\
				&-    &71.0 &54.7 &75.0 &64.2 &83.2 &72.6 \\
				\hline
				\multirow{2}*{TKP-D}&\checkmark     &66.3 &50.3 &74.2 &61.7 &79.3 &66.1 \\
				&-     &70.3 &55.0 &75.0 &63.1 &84.1 &72.9 \\
				\hline
		\end{tabular}}
	\end{center}
	\vspace{-15pt}
	\label{tab:bp2v}
\end{table}

\begin{table}[t]
	\centering
	\small
	\captionsetup{font={small}}
	\caption{Comparison with the method using pre-trained video model on the MARS dataset.}
	\vspace{-20pt}
	\begin{center}{
			\setlength{\tabcolsep}{2mm}
			\begin{tabular}{l |c c c }
				\hline
				Models  &baseline  &pre-trained &TKP \\
				\hline
				top-1  &67.1 &73.2 &\bfseries75.6 \\
				mAP    &55.5 &61.5 &\bfseries65.1 \\
				\hline
		    \end{tabular}}
	\end{center}
	\vspace{-15pt}
	\label{tab:pretrained}
\end{table}

\begin{table}[t]
	\centering
	\small
	\captionsetup{font={small}}
	\caption{Comparing the methods with different identification losses on the MARS dataset.}
	\vspace{-20pt}
	\begin{center}{
			\setlength{\tabcolsep}{1mm}
			\begin{tabular}{l |c c c}
				\hline
				Models    &I2V tri.  &Integrated tri. &baseline\\
				\hline
				top-1   &54.4 &59.1 &\bfseries67.1\\
				mAP     &42.6 &47.3 &\bfseries55.5\\
				\hline
		    \end{tabular}}
	\end{center}
	\vspace{-15pt}
	\label{tab:losses}
\end{table}

\noindent
{\bf Whether the non-local blocks are required?} 
In our framework, we use non-local blocks to model temporal relations between video frames.
To verify whether the non-local blocks are required or not, we remove the non-local blocks from the methods \textit{baseline} and \textit{TKP-F}.
And the video frame feature $f_{nt}$ in Eq.~\eqref{eq:tkpviafeature} is replaced by the video feature $v_n$ after temporal average pooling.
As shown in Table~\ref{tab:nonlocal}, when the non-local blocks are removed, \textit{TKP-F} still supasses \textit{baseline} by a reasonable margin.
But the performance and improvement without non-local blocks are both lower than those with non-local blocks. 
We argue that, compared with simple temporal average pooling, non-local blocks can model temporal information much better, which makes temporal knowledge propagation more effective as well.

\vspace{5pt}
\noindent
{\bf Whether TKP loss should propagate gradient to video representation network?}
As discussed in Section~\ref{sec:tkp}, enforcing TKP loss to propagate gradient to video representaiton network will degenerate video representations \wrt temporal knowledge.
To verify this, we add two additional experiments and the results are reported in Table~\ref{tab:bp2v}.
It can be seen that, when the gradient of TKP loss is propagated to video network, \textit{TKP-F} and \textit{TKP-D} can still increase the performance of I2I and I2V Re-ID by a considerable margin.
But these two methods consistently gain lower V2V performance.
If the back propagation to video network is banned, all I2I, I2V and V2V results can be further improved.

\vspace{5pt}
\noindent
{\bf Whether using a pre-trained video model is beneficial for the convergence of networks?}
Our method aims to solve I2V matching, but the pre-trained video model supervised by V2V loss function may not be optimal for I2V matching.
To verify this, we add an experiment which uses a pre-trained video model to perform knowledge propagation.
As shown in Table~\ref{tab:pretrained}, though the method \textit{pre-trained} outperforms \textit{baseline}, it is inferior to \textit{TKP} which learns two networks simultaneously.

\vspace{5pt}
\noindent
{\bf The influence of different triplet losses.}
To explore this, we perform the experiments with different kinds of triplet losses.
As shown in Table~\ref{tab:losses}, the method \textit{Integrated tri.} using integrated triplet loss surpasses \textit{I2V tri.} which only uses I2V triplet loss.
With additional classification loss, \textit{baseline} outperforms these two methods.

\begin{figure}[t]
	\centering
	\includegraphics[width = 1\columnwidth]{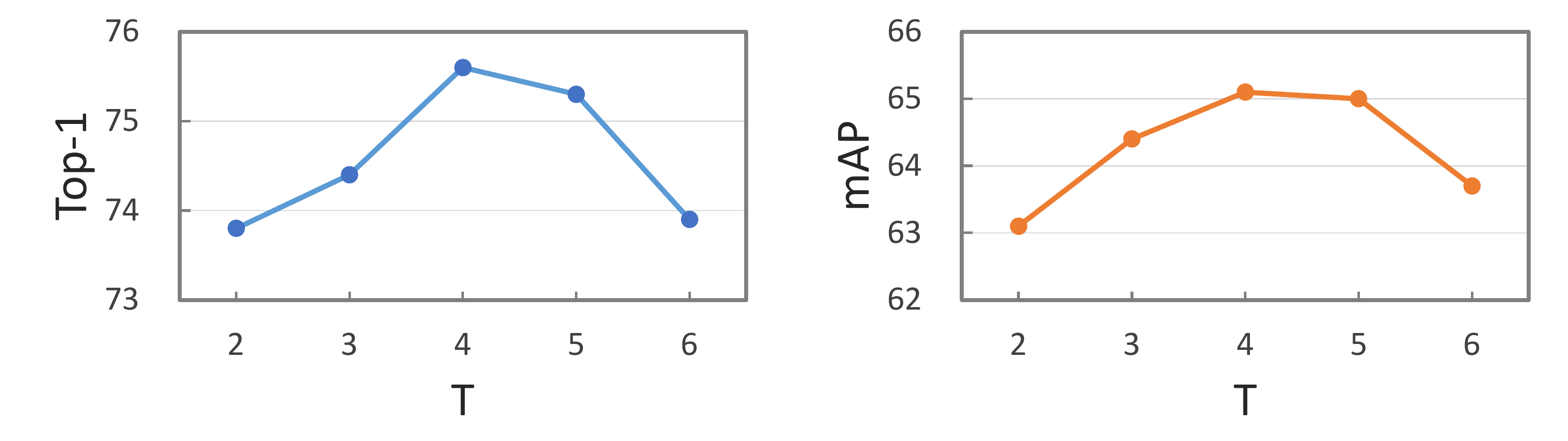}
	\vspace{-25pt}
	\captionsetup{font={small}}
	\caption{The results with different $T$ on the MARS dataset.}
	\vspace{-15pt}
	\label{fig:T}
\end{figure}

\vspace{5pt}
\noindent
{\bf The influence of the video clip size $T$.}
By varying $T$, we show the experimental results in Figure~\ref{fig:T}.
It can be seen that the best top-1 and mAP are consistently achieved when $T$ is $4$.

\section{Conclusion}

In this paper, we propose a novel TKP method for I2V Re-ID. 
TKP can transfer temporal knowledge from video representation network to image representation network.
With the transferred temporal knowledge, the robustness of the image features can be improved and the information asymmetry between image and video features can also be alleviated.
Extensive experiments demonstrate the effectiveness of our method and the results on two widely used datasets significantly surpass start-of-the-art performance.

\vspace{3pt}
\noindent
{\bf Acknowledgement} This work is partially supported by National Key R\&D Program of China (No.2017YFA0700800), Natural Science Foundation of China (NSFC): 61876171 and 61572465.

{\small
\bibliographystyle{ieee_fullname}
\bibliography{egbib}
}

\end{document}